\documentclass[11pt]{article}

\usepackage[inline]{enumitem}
\usepackage{pbox}
\usepackage{balance}
\newlist{inlinelist}{enumerate*}{1}
\setlist*[inlinelist,1]{label=\roman*),itemjoin={{, }},itemjoin*={{, and }}}
\usepackage{url}
\usepackage{multirow}
\usepackage{multicol}
\usepackage{booktabs}
\usepackage{tikz}
\usepackage{xcolor}
\usepackage{lipsum}
\usepackage{acronym}
\usepackage{subcaption}
\usepackage[inline]{enumitem}
\usepackage{xspace}
\usepackage{arydshln}
\usepackage{array}
\usepackage{adjustbox}
\usepackage{newfloat}
\DeclareFloatingEnvironment[fileext=lop, listname={List of prompts}, 
                            name=Prompt, placement=h]{prompt}
\usepackage{listings}
\usepackage{xcolor}
\lstdefinelanguage{PolicyPrompt}{
  literate=
    {<think>}{{{\color{green!60!black}<think>}}}{7}
    {</think>}{{{\color{green!60!black}</think>}}}{8}
    {<search>}{{{\color{orange!80!black}<search>}}}{8}
    {</search>}{{{\color{orange!80!black}</search>}}}{9}
    {<information>}{{{\color{blue!80!black}<information>}}}{13}
    {</information>}{{{\color{blue!80!black}</information>}}}{14}
    {<answer>}{{{\color{red!70!black}<answer>}}}{8}
    {</answer>}{{{\color{red!70!black}</answer>}}}{9}
    {context\_block}{{{\color{red!90!black}context\_block}}}{13}
    {last\_user\_utterance}{{{\color{red!90!black}last\_user\_utterance}}}{19}
}

\usepackage{latex/acl}
\usepackage{times}
\usepackage{latexsym}
\usepackage{amsmath}
\usepackage{amssymb}
\usepackage[T1]{fontenc}
\usepackage[utf8]{inputenc}
\usepackage{microtype}
\usepackage{inconsolata}
\usepackage{graphicx,xcolor}
\usepackage{booktabs}
\usepackage{pifont}

\newcommand{\tthink}{\texttt{\textcolor{green!60!black}{<think>}}}
\newcommand{\tethink}{\texttt{\textcolor{green!60!black}{</think>}}}
\newcommand{\tsearch}{\texttt{\textcolor{orange!80!black}{<search>}}}
\newcommand{\tesearch}{\texttt{\textcolor{orange!80!black}{</search>}}}
\newcommand{\tinfo}{\texttt{\textcolor{blue!80!black}{<information>}}}
\newcommand{\teinfo}{\texttt{\textcolor{blue!80!black}{</information>}}}
\newcommand{\tanswer}{\texttt{\textcolor{red!70!black}{<answer>}}}
\newcommand{\teanswer}{\texttt{\textcolor{red!70!black}{</answer>}}}
\newcommand{\tctx}{\texttt{\textcolor{red!90!black}{context\_block}}}
\newcommand{\tlast}{\texttt{\textcolor{red!90!black}{last\_user\_utterance}}}

\usepackage{pgfplots}
\pgfplotsset{compat=1.18}

\newcommand{\header}[1]{\vspace{2mm}\noindent\textbf{#1}}

\acrodef{CS}{Conversational Search}
\acrodef{CSA}{Conversational Search Agent}
\acrodef{PTKB}{Personal Text Knowledge Base}
\acrodef{TREC}{TExt Retrieval Conference}
\acrodef{iKAT}{Interactive Knowledge Assistance Track}
\acrodef{CAsT}{Conversational Assistance Track}
\acrodef{NIST}{National Institute of Standards and Technology}
\acrodef{LLM}{Large Language Model}
\acrodef{LSR}{Learned Sparse Retrieval}

\acrodef{IR}{Information Retrieval}
\acrodef{NLP}{Natural Language Processing}
\acrodef{PEFT}{Parameter-Efficient Fine-Tuning}
\acrodef{ICL}{In-Context Learning}
\acrodef{LoRA}{Low-Rank Adaptation}

\acrodef{CQR}{Conversational Query Rewriting}

\acrodef{MSE}{Mean Square Error}

\newcommand{\chatrone}{ChatR1\xspace}

\newcolumntype{P}[1]{>{\centering\arraybackslash}p{#1}}

\author{
  Simon Lupart \\
  University of Amsterdam \\
  Amsterdam, The Netherlands \\
  \texttt{s.c.lupart@uva.nl} 
  \And
  Mohammad Aliannejadi \\
  University of Amsterdam \\
  Amsterdam, 
  The Netherlands \\
  \texttt{m.aliannejadi@uva.nl} 
  \And
  Evangelos Kanoulas \\
  University of Amsterdam \\
  Amsterdam, The Netherlands \\
  \texttt{e.kanoulas@uva.nl} 
}

\title{ChatR1: Reinforcement Learning for Conversational Reasoning and Retrieval Augmented Question Answering}

\begin{document}

\maketitle

\begin{abstract}
We present \chatrone, a reasoning framework based on reinforcement learning (RL) for conversational question answering (CQA). Reasoning plays an important role in CQA, where user intent evolves across dialogue turns, and utterances are often underspecified, requiring contextual interpretation, query reformulation, and dynamic coordination between retrieval and generation. Unlike static `rewrite, retrieve, and generate' pipelines, \chatrone interleaves search and reasoning across turns, enabling exploratory and adaptive behaviors learned through RL.
To address the challenge of sparse and delayed rewards in RL, we propose an intent-aware reward that provides turn-level feedback by aligning retrieval and reasoning with evolving user goals. 
\chatrone demonstrates strong performance on both 3B and 7B model backbones, outperforming competitive models on five CQA datasets, measured by different metrics (F1, BERTScore, and LLM-as-judge). We include a diverse set of CQA datasets to cover topic shifts, evolving intents, mixed-initiative dialogues, and multi-document grounding, testing \chatrone's performance from various aspects.
Ablation studies confirm the effectiveness of the intent-aware reward. Our analyses further reveal diverse reasoning trajectories and effective use of the search tool. ChatR1 also generalizes robustly across domains, demonstrating that RL-based reasoning enables more flexible and context-aware behavior than static CQA pipelines.\footnotemark
\end{abstract}




\section{Introduction}
\label{sec:introduction}

Reasoning models for information seeking have made rapid progress with the development of large language models (LLMs) and reinforcement learning (RL)~\cite{guo2025deepseek}. These systems are increasingly capable of fulfilling complex information needs, requiring multi-step reasoning, query decomposition, and tool use~\cite{jin2025search,chen2025learning,song2025r1}.\footnotetext{Code, models and datasets available on HuggingFace and GitHub at \url{https://github.com/SimonLupart/ChatR1}.} 
By integrating search engines as external tools, RL-trained models move beyond static, supervised fine-tuned retrieval-augmented pipelines, instead, learning dynamic retrieval and reasoning from delayed rewards~\cite{kaelbling1996rlsurvey,chu2025sft}. 
However, current RL reasoning frameworks remain limited to single-turn interactions with users, assuming explicit and isolated user questions~\cite{li2025reinforcement}, whereas commercial systems are increasingly moving toward multi-turn conversational search.\footnote{e.g., Perplexity.ai:~\url{https://www.perplexity.ai/}, OpenAI:~\url{https://chatgpt.com/features/deep-research/}.} Extending RL reasoning models to conversational scenarios introduces new reasoning complexities, as user intent evolves across turns and must be inferred from dialogue context~\cite{10.1145/3477495.3532678/convIR,surveyCS}.

\begin{figure}[t]
    \centering
    \includegraphics[width=\linewidth]{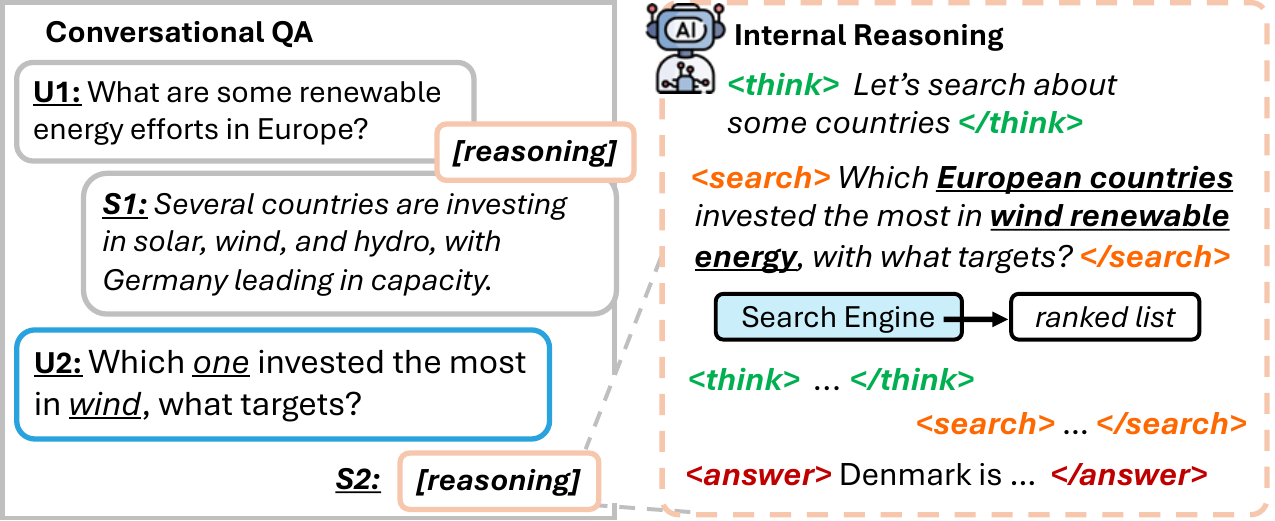}
    \caption{Reasoning in CQA requires the system to understand the user intent. The system resolves anaphoras and ellipses in the last user utterance ($\textbf{U2}$) during its internal reasoning, and formulates a search query, before producing the final answer.}
    \label{fig:reasoncqa}
\end{figure}

Reasoning is particularly valuable for conversational question answering (CQA) as dialogues contain evolving and underspecified information needs that require reference resolution, query reformulation, and coordination between retrieval and answer generation~\cite{elgohary-etal-2019-unpack}. As illustrated in Figure~\ref{fig:reasoncqa}, user utterances depend on earlier context (``one'' referring to ``European countries,'' ``wind'' to ``wind energy,'' along with a comparison of previously cited countries). Systems must interpret context as an additional reasoning step~\cite{theoreticalframeCS}. 
CQA also spans heterogeneous datasets that capture distinct user interaction patterns, ranging from topic transitions and mixed-initiative to task-oriented exchanges~\cite{wu2023inscit,adlakha-etal-2022-topiocqa}.
This diversity poses a unique generalization challenge, as models trained in one conversational domain often struggle to adapt to others~\cite{surveyCS}.
As discussed by \citet{chu2025sft}, while supervised fine-tuning (SFT) captures behaviors from demonstration data, reasoning from RL would enable broader generalization to conversational scenarios not explicitly seen during training. Despite the benefits of reasoning models for CQA, existing work relies exclusively on SFT~\cite{acikgoz-etal-2025-single,mo-etal-2025-uniconv,xu2024chatqav2}, leaving RL-based conversational reasoning largely unexplored.

In this work, we propose an \emph{RL framework for reasoning in CQA}, enabling models to learn retrieval and grounding strategies dynamically from interaction feedback. 
This approach differs from static retrieval pipelines by optimizing end-to-end reasoning behavior across multi-turn conversations.

While RL offers a promising framework for adaptive reasoning, its application to multi-turn CQA remains challenging because outcome-based rewards provide only delayed feedback with little guidance on which intermediate decisions, such as context interpretation or query formulation, contribute to success~\cite{2018suttonrl}.
This reward sparsity amplifies instability and hinders learning in conversational settings, where reasoning steps are interdependent and user intent evolves continuously~\cite{li2025reinforcement,zhang2025survey}.
We thus introduce an \emph{intent-aware reward} that provides turn-level supervision by leveraging canonical user intent annotations from CQA datasets. In our experiments, we compare this signal with an alternative retrieval-based reward (e.g., document coverage), showing that our intent-aware shaping better aligns intermediate decisions with evolving goals and complements final rewards.

Our results show that RL alone offers limited benefits over SFT; however, introducing our intent-aware reward enables adapting RL reasoning to CQA. With this design, our proposed model, \textbf{\chatrone}, achieves substantial performance gains in both in-domain and out-of-domain settings, surpassing SFT baselines and highlighting the importance of reward shaping for multi-turn reasoning.

\noindent{Our main contributions are as follows:}
\begin{itemize}
\item We introduce \chatrone, an RL-based reasoning model for CQA. \chatrone optimizes multi-turn retrieval and generation end-to-end, learning dynamic behavior rather than a static pipeline. Extensive experiments demonstrate the performance gains and generalization capabilities of \chatrone across different conversational complexities.
\item We propose an \emph{intent-aware reward} tailored for CQA, which reduces the sparsity of the reward signal by aligning retrieval behavior with evolving user intent across turns. Ablations show the benefit of this reward compared to other intermediate rewards.
\item Our analysis further reveals that \chatrone generates diverse reasoning paths, reflected in their lengths, and generalizes robustly across conversational domains. We also evaluate the performance of the retrieval and how it impacts generation performance to better understand the search usage of \chatrone.
\end{itemize}

\section{Related Work}
\label{sec:related}

\begin{figure*}[t!]
    \centering
    \includegraphics[width=\linewidth]{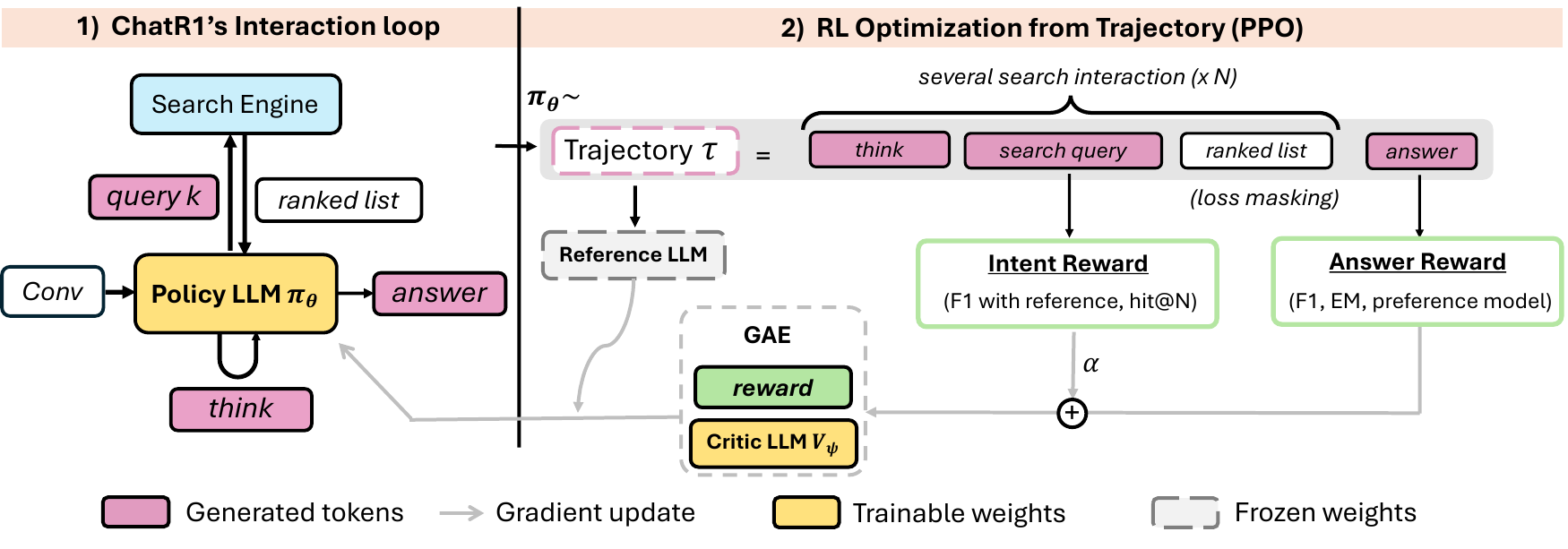}
    \caption{Overview of \chatrone's interaction loop: the policy model interleaves reasoning, search, and answering while interacting with a search environment. (Right) Full trajectories are sampled from the policy LLM and consist of all reasoning, search, and answer steps, with rewards assigned to search queries (intent-aware reward) and the final answer. These signals jointly train the policy under PPO optimization.}
    \label{fig:chatr1}
\end{figure*}

\header{Conversational question answering}
is the task of responding to user queries within multi-turn dialogues, with dependence on previous turns and often grounding from external knowledge~\cite{choi-2018-quac}. Retrieval-augmented generation (RAG) has proven particularly effective in CQA, reducing hallucinations and providing factual grounding in answers~\cite{mo-etal-2025-uniconv,chatqa}.
Proprietary systems such as Perplexity Sonar, Gemini Grounding, GPT Search and Claude have also demonstrated promising search and generation capabilities~\cite{miroyan2025search}. However, these industrial models remain closed-source, and it is unclear whether their behavior results from explicit reasoning or large-scale supervised pipelines composed of multiple static components.
While academic research has advanced CQA with retrieval and generation, most approaches still rely on static retrieval pipelines and lack explicit reasoning mechanisms for deciding when and how to search~\cite{xu2024chatqav2}. 
Similarly, in conversational search (CS), community evaluations emphasize user intent understanding and more complex user interactions~\cite{10.1145/3397271.3401206/cast19,mohikat23,mohikat24,gohsen2025user}. A common strategy is to model user information needs explicitly with \emph{query rewriting}, to align underspecified user questions with user intent~\cite{yu2021few,lupart2025disco,lupart2025investigating,abbasiantaeb-etal-2026-generating}. RL has also been applied to query rewriting~\cite{Wu2021CONQRRCQ,zhu2025convsearchr1}; both works design rewards over retrieved passages to improve query rewriting. However, these approaches do not extend to answer generation, leaving reasoning in CQA largely unexplored. We aim to fill this gap with a reasoning model that can understand user intent and interact with both the user and a search engine.

\header{Reasoning and QA with RL.} Recent work has applied RL to train LLMs as reasoning agents for information seeking~\cite{zhang2025survey,li2025reinforcement}. Approaches such as Search-R1~\cite{jin2025search}, R1-Searcher~\cite{song2025r1}, ReSearch~\cite{chen2025learning}, DeepResearcher~\cite{zheng2025deepresearcher} optimize policies that decompose complex questions, issue multiple queries, and integrate evidence, achieving strong results on knowledge-intensive QA. While early studies~\cite{jin2025search} relied solely on sparse trajectory-level rewards, some later work such as SearchR1++~\cite{jin2025empirical} and StepSearch~\cite{wang2025stepsearch} focused on the credit assignment problem and reward sparsity by introducing step-level rewards on intermediate reasoning or retrieval quality. In contrast, our approach extends this line of work by defining an intermediate reward tailored to CQA, capturing user intent at each dialogue turn to provide finergrained learning signals.

In parallel, advances in agentic tool use have complemented RL-based reasoning, enabling models to plan and invoke external APIs or search engines~\cite{toolformer2023,singh2025agentic}. Among these, CALM~\cite{acikgoz-etal-2025-single} extends reasoning to multi-turn dialogues, interleaving tool calls with user interactions while emphasizing the challenge of maintaining coherent user intent across turns. However, CALM remains trained through SFT, relying on demonstration data; extending it to RL could enable better exploration of multi-turn reasoning strategies.

\section{Methodology}
\label{sec:method}

In this section, we introduce the CQA notation before presenting \chatrone, the RL objective, and our intent reward mechanism. 

\subsection{Notation and Problem Setup}
We consider a dataset $\mathcal{D}$ of user--system conversations, each composed of multiple turns, and a collection of passages $\mathcal{C}$. 
At each turn, the system receives the conversation history $\mathcal{H}$ (all previous utterances) and the current user query $q$.
The CQA task is to generate an answer $y$ to $q$, leveraging context from $\mathcal{H}$ and grounding in $\mathcal{C}$.
We further define the user intent via rewritten queries $q^{rw}$, obtained from human annotations, that resolve contextual references and ambiguities in $q$.
These rewrites are available only during training as supervision.

\subsection{ChatR1 RL Objective}
Figure~\ref{fig:chatr1} illustrates the main components of \chatrone, along with the trajectory and reward structure. \chatrone is a policy model $\pi_\theta$ that, at each turn, generates a trajectory $\tau$. 
The trajectory consists of a reasoning trace including thinking, intermediate search queries $Q = \{q^k\}_{k=1}^K$ issued to the search engine $\mathcal{R}$, and retrieved passages, followed by the final answer $y$. The policy model's instructions define all special tokens linked to the external tools. In particular, the \texttt{<search>} token triggers the retriever, and the retrieved documents are then added to the trajectory. The full instruction of \chatrone is provided in Table~\ref{tab:prompts1}. 

\header{Objective.} The objective of \chatrone is to maximize the expected reward given the conversation history, the last user utterance, and the search engine, while regularizing the policy toward the reference policy:
\begin{flalign}
\mathcal{J}(\theta)
=
\mathbb{E}&_{(q, \mathcal{H}) \sim \mathcal{D},\
\tau \sim \pi_\theta(\cdot|q, \mathcal{H}; \mathcal{R})}
\\
&\quad\,R(\tau)\,
-\beta\, D_{\mathrm{KL}}\!\big(\pi_\theta \,\Vert\, \pi_{\mathrm{ref}}\big)~.
\nonumber &
\end{flalign}

\header{Policy optimization.}
We optimize this objective using Proximal Policy Optimization (PPO)~\cite{schulman2017proximal}, a policy-gradient algorithm designed to stabilize updates. 
PPO maximizes the clipped surrogate objective:
\begin{align}
&\mathcal{L}^{\mathrm{PPO}}(\theta)
=
\mathbb{E}_{(q,\mathcal{H}; \mathcal{R}; i) \sim \mu}
\\
&\quad\Big[
\min\big(
\rho_i(\theta)\,\hat{A}_i,\,
\operatorname{clip}(\rho_i(\theta), 1-\epsilon, 1+\epsilon)\,\hat{A}_i
\big)
\Big]\nonumber~,
\end{align}
\noindent
where the expectation under $\mu$ averages over conversation examples from the dataset, trajectories sampled from the policy, and generated token positions, following previous works on token loss masking~\cite{jin2025search}. We define the ratio $\rho$ as the probability ratio between the new and old policies.
\begin{equation}
\rho_i(\theta)
=
\frac{\pi_\theta(\tau_i \mid \mathcal{H}, q, \tau_{<i}; \mathcal{R})}
     {\pi_{\theta_{\mathrm{old}}}(\tau_i \mid \mathcal{H}, q, \tau_{<i}; \mathcal{R})}~.
\end{equation}
\noindent
We also represent $\hat{A}_i$, the estimated advantage at token position $i$, which quantifies how much better the chosen token $\tau_i$ performs compared to the expected value predicted by the critic. 
\begin{equation}
\hat{A}_i
=
\delta_i
+ (\gamma \lambda)\, \delta_{i+1}
+ \cdots
+ (\gamma \lambda)^{T-i-1}\, \delta_{T-1}~,
\end{equation}
\begin{equation}
\delta_i = \gamma\, V_\psi(\tau_{i+1}) - V_\psi(\tau_i)~,
\end{equation}
with $V_\psi(\tau_T) = R(\tau)$ the final reward for the last token of the trajectory, and $V_\psi$ the surrogate critic model.
This ensures that rewards are propagated across all generated tokens, allowing the model to assign credit throughout the reasoning and answer generation process. $V_\psi$ is optimized with Generalized Advantage Estimation (GAE)~\cite{schulman2015high} following existing literature.

\begin{table}[t]
\centering
    \begin{lstlisting}[language=PolicyPrompt]
You are a helpful assistant tasked with answering a user query. Your primary goal is to generate a complete and informative answer.
If the query is ambiguous or refers to earlier context (e.g., pronouns or ellipsis), use the conversation history provided below to resolve it.
- Always perform your reasoning inside <think>...</think>.
- If external information is needed, use <search>your query</search>. 
- Retrieved documents will appear between <information>...</information>.
- You may issue multiple search queries if needed.
- Once you have enough information, provide a complete answer within <answer>...</answer>.

Conversation context: {context_block}
User query: {last_user_utterance}
    \end{lstlisting}
  \caption{Instructions for the \chatrone policy LLM.}
  \label{tab:prompts1}
\end{table}

\subsection{Reward Modeling}
We design a composite reward function that captures both the quality of the final answer and the understanding of user intent throughout the trajectory. 
For each trajectory $\tau$, composed of a sequence of issued queries $Q = \{q^{1}, \ldots, q^{K}\}$ and final answer $y$, the total reward is defined as:
\begin{equation}
R(\tau) = R_{\text{answer}}(y) + \alpha \ R_{\text{intent}}(Q)~,
\end{equation}
where $\alpha$ balances the contribution of answer quality and intent understanding.

\header{Answer reward.}
The answer reward $R_{\text{answer}}$ evaluates how well the final system response aligns with the gold reference $y^\ast$. 
Following prior work~\cite{jin2025search,song2025r1}, we adopt a word-level F1 score to capture semantic overlap:
\begin{equation}
R_{\text{answer}}(y) = \mathrm{F1}(y, y^\ast)~,
\end{equation}
where $\mathrm{F1}(a,b) = \tfrac{2 \, \mathrm{P}(a,b) \, \mathrm{R}(a,b)}{\mathrm{P}(a,b) + \mathrm{R}(a,b)}$, computed from word-level precision and recall.
This reward is assigned only at the end of the trajectory, encouraging coherent intermediate reasoning steps that ultimately improve the final answer.

\header{Intent reward.}
The intent reward $R_{\text{intent}}$ provides additional feedback for the quality of intermediate search queries.
It measures how well any issued query $q^k$ captures the user intent, expressed by the human-authored rewrite $q^{rw}$:
\begin{equation}
R_{\text{intent}}(Q) = \max_{q^k \in Q} \, \mathrm{F1}(q^k, q^{rw})~.
\end{equation}
Taking the maximum allows for exploration while ensuring that at least one search query aligns with the user intent.
Compared to previous overlap-based query rewards such as StepSearcher~\cite{wang2025stepsearch}, our proposed reward explicitly targets user-intent alignment in the CQA setting, offering a denser, retrieval-agnostic supervision signal that remains informative even when retrieval or final answering fails. 

As an alternative reward, we also test a simpler retrieval-based variant that rewards retrieved relevant passages from the top-$N$ of any of the issued search queries~\cite{jin2025empirical}.

\section{Experiments}
\label{sec:exps}

\subsection{Experimental Setup}
\label{sec:expset}

\header{Dataset statistics.} We trained and compared \chatrone on five CQA datasets, containing multi-turn dialogues with evolving user intent.
Table~\ref{tab:dataset_complex} lists the train and test splits, as well as the main challenges of each dataset.
TopiOCQA~\cite{adlakha-etal-2022-topiocqa} contains open-domain conversations where questions often involve topic shifts, requiring models to retrieve and reason over new information as the dialogue progresses. QReCC~\cite{anantha-etal-2021-open-qrecc} is a large-scale conversational dataset on a large web corpus, paired with reformulations and gold answers. INSCIT~\cite{wu2023inscit} focuses on information-seeking dialogues with mixed initiative (e.g., clarification questions). This requires models to understand user feedback and use it accordingly. FaithDial~\cite{dziri-etal-2022-faithdial} is a conversational QA dataset, as an improved version of WoW~\cite{dinan2018wizard}, with better faithfulness and grounding to passages of the collection.
Finally, MultiDoc2Dial~\cite{feng-etal-2021-multidoc2dial} is a goal-oriented information-seeking dataset grounded in multiple governmental documents. More details can be found in Appendix~\ref{sec:appendix:stat}.

\begin{table}[t!]
    \centering
    \small
    \adjustbox{max width=\columnwidth}{
    \begin{tabular}{lcc}
    \toprule
    {Dataset} & \textbf{\#Turns} & \textbf{Challenges} \\
    \midrule
    TopiOCQA & 45k & Topic shifts, \\
    ~\cite{adlakha-etal-2022-topiocqa} & 2.5k & evolving intent \\
    \midrule
    QReCC & 63k & Large-scale corpus, \\
    ~\cite{anantha-etal-2021-open-qrecc} & 16k & query reformulation \\
    \midrule
    INSCIT & 1.8k & Mixed initiatives, \\
    ~\cite{wu2023inscit} & 3.3k & open intent \\
    \midrule
    MultiDoc2Dial & 18k & Multi-doc grounding, \\
    ~\cite{feng-etal-2021-multidoc2dial} & 3.3k & in-domain reasoning \\
    \midrule
    FaithDial & 18k & Faithfulness, \\
    ~\cite{dziri-etal-2022-faithdial} & 3.5k & hallucination control \\
    \bottomrule
    \end{tabular}
    }
    \caption{Challenges of respective datasets with numbers of turns in train and test splits.}
    \label{tab:dataset_complex}
\end{table}


\header{Evaluation metrics.} We report the results of the generation task in terms of F1, BERTScore, and using LLM-as-judge, all with respect to a human gold reference answer. The prompt for the LLM-as-judge metric is provided in Appendix~\ref{sec:appendix:prompt}, with \texttt{gpt-4o-mini}, following established practice~\cite{song2025r1,rau-etal-2024-bergen}.
For retrieval, we report standard IR metrics, including Normalized Discounted Cumulative Gain (nDCG), recall (R), and mean reciprocal rank (MRR). We refer to hit@N as the proportion of queries on which a relevant passage is within the top N of the retrieved list, following previous works~\cite{ndcg,mohikat24}.

\header{Baselines.}
We compare \chatrone with a wide range of competitive baselines: zero-shot, supervised fine-tuned, and RL models. Direct inference (\textbf{DI}) with GPT-3.5, Claude, and Qwen2.5-3B-Instruct models generates answers from internal model knowledge, with the variant using chain-of-thought (\textbf{CoT}). \textbf{IRCoT} similarly uses chains of thought in combination with a search engine using the prompt of \chatrone in zero-shot. Retrieval augmented generation (\textbf{RAG}) uses query rewrite from LLM to do conversational retrieval, and then uses the LLM to generate the final answer based on retrieved passages. All of these models take advantage of their long input context windows to understand the history and the last user utterance.

\begin{table*}[t!]
    \centering
    \small
    \adjustbox{max width=\textwidth}{
    \begin{tabular}{l l l *{5}{cc}}
        \toprule
        \multirow{2}{*}{{Method}} & \multirow{2}{*}{\textbf{RAG}}  & \multirow{2}{*}{\textbf{LLM}} 
        & \multicolumn{2}{c}{\textbf{TopiOCQA}} 
        & \multicolumn{2}{c}{\textbf{QReCC}} 
        & \multicolumn{2}{c}{\textbf{INSCIT}} 
        & \multicolumn{2}{c}{\textbf{MD2Dial}}
        & \multicolumn{2}{c}{\textbf{FaithDial}} \\
        \cmidrule(lr){4-5} \cmidrule(lr){6-7} \cmidrule(lr){8-9} \cmidrule(lr){10-11} \cmidrule(lr){12-13}
         & & & \textbf{F1} & \textbf{Bert} & \textbf{F1} & \textbf{Bert} & \textbf{F1} & \textbf{Bert} & \textbf{F1} & \textbf{Bert} & \textbf{F1} & \textbf{Bert} \\
        \midrule
        \multicolumn{13}{l}{\textit{Zero-Shot}} \\
        \midrule
        Qwen-Instr. (DI) & No & Qwen-3b & 6.7 & 57.3 & 13.3 & 55.3 & 17.9 & 58.1 & 13.2 & 64.4 & 10.9 & 64.8 \\
        Qwen-Instr. (CoT) & No & Qwen-3b   & 6.7 & 58.6 & 12.8 & 58.5 & 16.4 & 62.4 & 10.5 & 63.6 & 9.6 & 64.8 \\
        IRCoT & RAG & Qwen-3b & 8.9 & 61.0 & 13.1 & 55.6 & 20.4 & 67.3 & 13.3 & 67.5 & 9.6 & 64.7 \\
        Qwen-Instr. (RAG) & RAG & Qwen-3b & 8.8 & 54.7 & 15.5 & 64.5 & 13.0 & 49.3 & 18.8 & 75.1 & 12.3 & 73.8 \\
        ChatGPT (DI) & No & GPT-3.5 & 25.5 & 77.5 & 22.6 & 75.6 & 22.8 & 81.1 & 21.6 & 81.7 & 12.9 & 80.9 \\
        Claude (DI) & No & Claude & 27.2 & -- & 25.0 & -- & 27.0 & -- & -- & -- & -- & --  \\
        \midrule
        \multicolumn{13}{l}{\textit{Supervised Fine-tuning or RL Training}} \\
        \midrule
        conv-ANCE +Mis. & RAG & Mistral 7b & 27.2 & -- & 25.9 & -- & 24.8 & -- & -- & -- & -- & -- \\
        ChatRetriever +Mis. & RAG & Mistral 7b & 28.3 & -- & 26.3 & -- & 30.3 & -- & -- & -- & -- & -- \\
        UniConv & RAG & Mistral 7b & \underline{29.6} & -- & 26.2 & -- & \textbf{33.2} & -- & -- & -- &11.6 & -- \\
        CoT R1 & No & Qwen-3b & 12.5 & 70.3 & 17.7 & 72.6 & 24.1 & 84.0 & 18.0 & 80.2 &14.5 & 80.9 \\
        SFT & No & Qwen-3b   & 18.0 & 78.8 & 23.3 & \underline{80.0} & 16.9 & 56.9 & 25.4 & \underline{84.2} & \underline{18.6} & 83.8 \\
        QR Search R1  & RAG & Qwen-3b  & 20.1 & 72.7 & 20.4 & 79.6 & 27.5 & 84.0 & 23.1 & 82.1 & 14.4 & 82.2 \\
        \midrule
        \chatrone (w/o $\mathrm{R_{int.}}$) & RAG & Qwen-3b & 24.4 & 73.0 & 27.0 & 78.5 & 31.3 & 84.4 &\underline{26.4} & 77.4 &15.5 & 81.3 \\
        \chatrone-3b & RAG & Qwen-3b &29.4$^\dagger$ & \textbf{80.9}$^\dagger$ &\underline{28.0}$^\dagger$ & 79.2$^\dagger$ & \textbf{33.2}$^\dagger$ & \textbf{85.5}$^\dagger$ &26.0 & 83.1$^\dagger$ &\textbf{19.2}$^\dagger$ & \underline{84.0}$^\dagger$ \\
        \chatrone-7b & RAG & Qwen-7b &\textbf{30.6}$^\ddagger$  & \underline{79.5} &\textbf{31.0}$^\ddagger$ & \textbf{80.7}$^\ddagger$ & \underline{32.8} & \textbf{85.5} &\textbf{31.2}$^\ddagger$ & \textbf{84.5}$^\ddagger$ &18.1 & \textbf{84.8}$^\ddagger$ \\
        \bottomrule
    \end{tabular}
    }
    \caption{Conversational response generation comparison of \chatrone on five datasets, for zero-shot, SFT and RL baselines. Superscript $\dagger$ and $\ddagger$ are paired t-tests ($p<0.05$) between w/ vs. w/o intent reward, and between 3B and 7B variants, respectively. Best results in bold, second best underlined.}
    \label{tab:f1_bert_scores}
\end{table*}

We further compare \chatrone with supervised fine-tuned approaches, each trained on the same datasets as \chatrone. \textbf{CoT R1} is a chain-of-thought approach trained with an F1 RL-reward on the final answer, similar to \chatrone, without access to the retrieval tool. Supervised fine-tuning (\textbf{SFT}) is trained with a next-token prediction loss to generate the answer. \textbf{QR Search R1} uses query rewrite on the input conversation and then trains a Search R1 model on the query rewrites distribution.
Finally, we report the results of three RAG state-of-the-art methods, {\textbf{UniConv}}~\cite{mo-etal-2025-uniconv}, \textbf{conv-ANCE +Mistral-7B} and \textbf{ChatRetriever +Mistral-7B}~\cite{mao-etal-2024-chatretriever}. Those three models are fine-tuned on a large conversational dataset consisting of TopiOCQA and synthetic dialogs. They are also trained in two steps: first, the retrieval model with a contrastive loss, then the generation model with SFT. They also feature 7B encoder models for retrieval, while \chatrone retrieval is only 300M.

\header{Model and training details.}
We rely on \texttt{Qwen2.5-3B-Instruct} and its 7B variant as the base LLMs~\cite{yang2025qwen3}, fine-tuned on each five datasets. Retrieval model is \texttt{e5-base-v2} used in zero-shot. \chatrone and other baselines use the top-3 retrieved passages; we also limit the number of search calls to two, following previous works~\cite{jin2025search,jin2025empirical}. Query rewrites are human annotations; when not available, we used GPT-4.1 to generate them. The batch size of the policy model is 512, with a PPO micro batch size of 64, a max prompt length of 3500 tokens, and a learning rate of 1e-6 for the actor model. We rely on the GAE algorithm for the critic optimization. Both policy and critic models are initialized with the same LLM, with weights fine-tuned independently. We train for 500 steps, with a saved checkpoint every 50 training steps. We use the final checkpoint unless training rewards collapse. Additional details are provided in Appendix~\ref{sec:appendix:baselines}. All experiments are conducted on 4 H100 GPUs.

\subsection{Results}

\header{Conversational response generation.}
Table~\ref{tab:f1_bert_scores} presents the main results comparing \chatrone's generation performance with baselines. From the results, we observe that (i) \chatrone-3B achieves higher performance than closed-source commercial LLMs (ChatGPT, Claude) while using far fewer parameters, highlighting the effectiveness of RAG over parameter scaling.
\begin{figure}[t!]
    \centering
    \includegraphics[width=\linewidth]{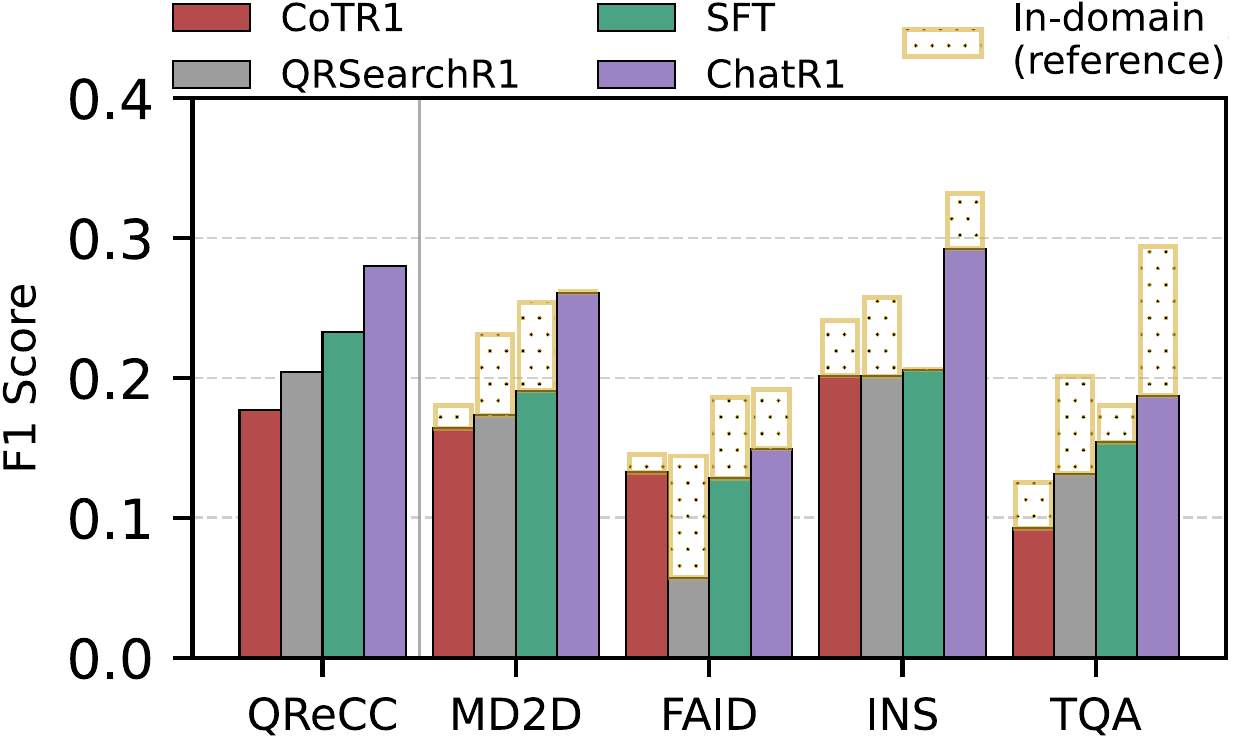}
    \caption{Out-of-domain transfer performance for models trained on QReCC, with in-domain reference.}
    \label{fig:ood_f1}
\end{figure}
(ii) \chatrone-3B surpasses all SFT- and RL-based 3B models in both F1 and BERTScore across most datasets. Moreover, \chatrone-3B matches or exceeds 7B supervised RAG baselines such as UniConv and ChatRetriever+Mistral on TopiOCQA, QReCC, and INSCIT. Notably, while UniConv and ChatRetriever rely on 7B retrieval backbones, \chatrone-3B employs a 300M retriever, highlighting its strong ability to leverage retrieval tools effectively. (iii) We further observe that scaling \chatrone from 3B to 7B parameters yields consistent performance gains, with an average improvement of 1.4 F1 points and 0.5 BERTScore. With this scaling, \chatrone-7B outperforms most compared baselines overall.

\header{Generalization.} To further assess the generalization ability of \chatrone, Figure~\ref{fig:ood_f1} reports its performance in out-of-domain settings, where models trained on QReCC are evaluated on the four remaining datasets. We also highlight in gold the reference in-domain performance. Interestingly, \chatrone-3B shows minimal loss on MultiDoc2Dial (only 0.2) and generally lower loss than other baselines across the remaining datasets. \chatrone shows slightly higher loss on TopiOCQA, likely due to the shorter answer lengths of the dataset responses, yet still surpasses all baselines. We also observe that \chatrone-3B still surpasses ChatGPT on three of the four datasets in zero-shot. This highlights how \chatrone learns to use the retrieval tool, rather than overfitting to domain-specific datasets.

\begin{figure}[t!]
    \centering
    \includegraphics[width=0.9\linewidth]{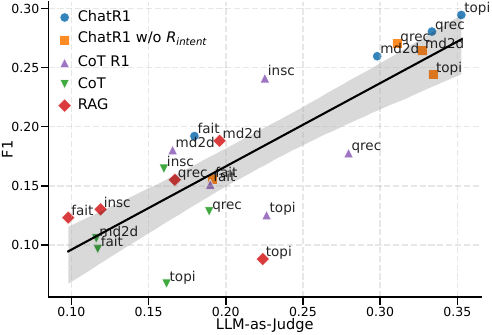}
    \caption{Correlation between using F1 score and LLM as judge on \chatrone and baselines across datasets.
    }
    \label{fig:llmjudge}
\end{figure}

\header{LLM evaluation.} Figure~\ref{fig:llmjudge} further illustrates \chatrone's performance when evaluated with LLM-as-judge. LLM-as-judge and F1 generally agree for in-domain evaluation with a Pearson's $r$ of 0.83, indicating agreement between both metrics. The results also demonstrate the clear benefit of retrieval and fine-tuning. Finally, many zero-shot approaches, even though they have low F1, still have reasonable LLM-as-judge performance, showing that fine-tuning enables the model to also learn the distribution of words in the answer, and the targeted format of gold answers.


\begin{figure}[t!]
    \centering
    \includegraphics[width=\linewidth]{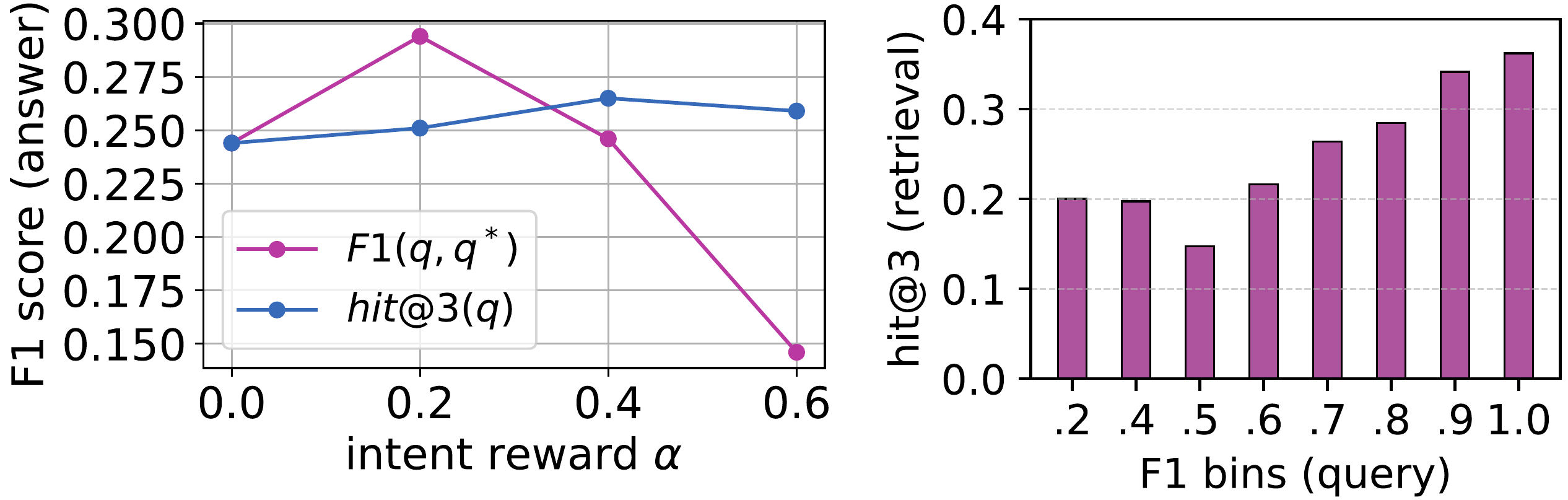}
    \caption{
    (Left) Analysis of intent rewards, contrasting passage-level rewards (hit@k) with query-level rewards (F1), and their impact on conversational QA response quality. (Right) Relationship between hit@k and query-level F1. Both on TopiOCQA with \chatrone-3B.}
    \label{fig:ablation}
\end{figure}

\begin{figure}[t!]
    \centering
    \includegraphics[width=\linewidth]{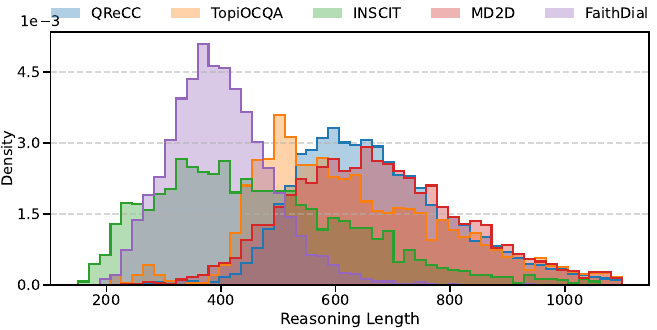}
    \caption{
    Reasoning length of \chatrone-7B when trained on the different datasets.}
    \label{fig:reasoninglen}
\end{figure}




\subsection{Analysis}

While the previous section discussed \chatrone's performance, this section presents an ablation study and analyses of reasoning length and retrieval quality to examine variability in model outputs and retrieval behavior.

\begin{table}[t!]
\centering
\small
\begin{tabular}{lccccc}
\toprule
ChatR1-3B & \textbf{TQA} & \textbf{QRC} & \textbf{INS} & \textbf{MD2D} & \textbf{FD} \\
\midrule
Jacc. overlap & 	0.52 & 0.25 & 0.12 & 0.29 & 0.57 \\
\bottomrule
\end{tabular}
\caption{Jaccard overlap between consecutive search queries used by ChatR1.}
\label{tab:jaccard_overlap}
\end{table}

\header{Reward ablations.}
We conduct ablations on the generation of \chatrone to study the effect of the intent reward. First, as shown in Table~\ref{tab:f1_bert_scores}, \chatrone-3B outperforms the base model without the intent reward by an average of 2.2 F1 points. We also explored the impact of the $\alpha$ parameters and reward in Figure~\ref{fig:ablation}. In particular, we observe that the intent reward on search queries directly $F1(q,q^*)$ leads to better performance compared to the reward on the retrieved passages with hit@3. This can be explained from several perspectives:
(i) F1 on search queries is less sparse, providing a stronger learning signal for PPO and more stable optimization with dense rewards~\cite{zhu2025convsearchr1,2018suttonrl};
(ii) it offers a direct signal on query formulation, independent of the search engine, thereby disentangling retrieval errors from formulation errors; and
(iii) passage relevance judgments have annotation gaps (i.e., some passages may contain valid answers but are not labeled as relevant), causing hit-based rewards to be incomplete. In contrast, query-level rewards directly assess the quality of the formulated query and are not affected by such labeling gaps.
From a requirements perspective, constructing complete passage-query relevance pairs is considerably more expensive than generating rewrites. Finally, we observe that the best performance is achieved with a reward ratio of 0.2/1.0 between retrieval and generation rewards.
As shown on the right side of the figure, $F1(q, q^*)$ is less sparse but still correlates with the hit metric.

\header{Reasoning paths.}
Figure~\ref{fig:reasoninglen} illustrates the diversity in reasoning length across the five evaluated datasets. We observe that MultiDoc2Dial and QReCC exhibit the longest reasoning traces, while FaithDial features comparatively shorter ones. This is expected, as MultiDoc2Dial involves multi-hop CQA, requiring the model to perform deeper reasoning and multiple retrieval steps to synthesize relevant information. In contrast, INSCIT shows a much more dispersed distribution, which can be attributed to its mixed-initiative nature; some turns demand substantial reasoning due to user-driven complexity, while others remain relatively simple. 

To further investigate whether the model tends to produce near-duplicate rewrites, we computed the Jaccard word-overlap between ChatR1's first and second queries when it generates $>2$ queries for a turn. From Table~\ref{tab:jaccard_overlap}, we observe some low similarity on INSCIT, MultiDoc2Dial, and QReCC with all values $<0.30$, but higher similarity on FaithDial and TopiOCQA. This shows that the model explores the collection, which is the targeted behavior.

\begin{table}[t]
    \centering
    \small
    \begin{tabular}{l*{4}{c}}
        \toprule
        \multirow{2}{*}{Model} 
        & \multicolumn{2}{c}{\textbf{TopiOCQA}} 
        & \multicolumn{2}{c}{\textbf{QReCC}} \\
        \cmidrule(lr){2-3} \cmidrule(lr){4-5}
        & \textbf{N@3} & \textbf{R@10} 
        & \textbf{N@3} & \textbf{R@10} \\
        \midrule
        \multicolumn{5}{l}{\textit{Query Rewrite}} \\
        \midrule
        Gold Rewrite  & 30.6 & 52.5 & 40.6 & 67.2 \\
        \midrule
        QuReTeC & 10.5 & 20.2 & 32.6  & 55.0 \\
        Qwen-3B QR  & 19.6 & 35.3 & 27.2 & 45.5 \\
        T5QR & 22.2 & 37.6 & 31.8 & 53.1 \\
        LLM4CS & \textbf{26.7} & 43.3 & \textbf{42.1} & \textbf{66.4} \\
        \midrule
        \multicolumn{5}{l}{\textit{Encoder-based Retrieval}} \\
        \midrule
        ConvDR        & \underline{26.4} & 43.5 & 35.7 & 58.2 \\
        \midrule
        \multicolumn{5}{c}{\textit{Retrieval-as-a-Tool}} \\
        \midrule
        IRCoT-3B & 18.7 & 35.6 & 31.2 & 52.8 \\
        ChatR1-3B  & 24.1 & \underline{43.7} & 36.4 & 60.2 \\
        ChatR1-7B  & \textbf{26.7} & \textbf{46.9} & \underline{37.0} & \underline{61.1}  \\
        \bottomrule
    \end{tabular}
    \caption{Retrieval-as-a-tool performance of \chatrone on TopiOCQA and QReCC datasets with dense retrieval.}
    \label{tab:retrieval_performance}
\end{table}

\header{Conversational search.}
Table~\ref{tab:retrieval_performance} highlights the retrieval performance of \chatrone, which leverages retrieval as a tool without relying on direct supervision. Despite this, \chatrone achieves results that are on par with state-of-the-art query rewriting and encoder-based methods. In particular, ChatR1-7B matches or even surpasses several supervised baselines across both TopiOCQA and QReCC, demonstrating the effectiveness of reasoning-driven retrieval emerging from interactive learning. The consistent gains from 3B to 7B also show that scaling enhances retrieval reasoning rather than merely memorization. Overall, these findings confirm that \chatrone can autonomously learn to perform retrieval effectively, approaching the performance of systems explicitly trained for query rewriting.

\begin{figure}[t!]
    \centering
    \includegraphics[width=\linewidth]{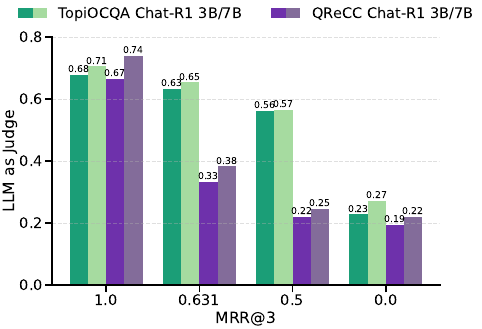}
    \caption{
    Retrieval quality effect on generation performance for \chatrone.}
    
    \label{fig:rag_retrieval}
\end{figure}

Figure~\ref{fig:rag_retrieval} further illustrates how retrieval quality impacts generation performance in \chatrone. The model exhibits a clear trend where higher retrieval accuracy translates to improved answer quality. Both 3B and 7B variants show consistent improvements across retrieval bins, confirming that \chatrone effectively adapts its reasoning process to the quality of retrieved evidence. The 7B model offers uniform gains across all bins, suggesting that scaling enhances overall robustness rather than compensating only for poor retrieval. Interestingly, the performance jump from low to mid retrieval quality is more pronounced on QReCC than on TopiOCQA. This is likely because QReCC, as a web-based dataset with longer passages, requires the model to process and verify more context within each retrieved document, making retrieval accuracy more critical to final generation quality.

\begin{figure}[t!]
    \centering
    \includegraphics[width=\linewidth]{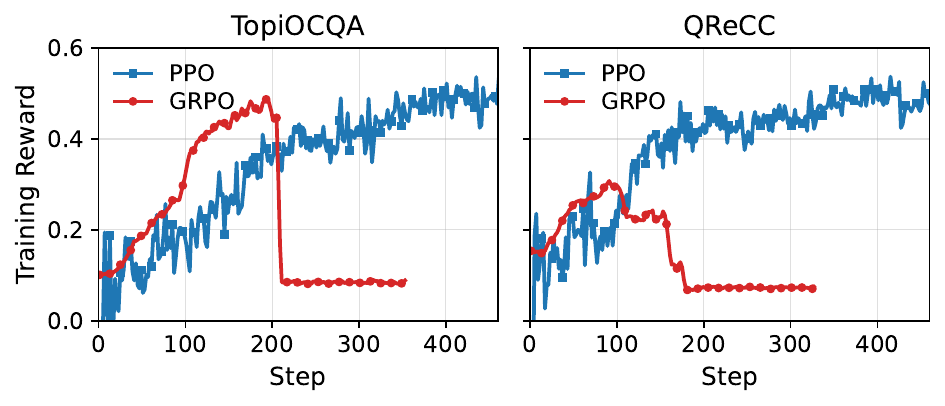}
    \caption{
    Training of \chatrone-3B with PPO and GRPO.}
    \label{fig:train}
\end{figure}

\begin{table}[t]
\centering
\small
\adjustbox{max width=0.49\textwidth}{
\begin{tabular}{lccccc}
\toprule
{Method / Dataset} & \textbf{TQA} & \textbf{QRC} & \textbf{INS} & \textbf{MD2D} & \textbf{FD} \\
\midrule
ChatR1 +PPO  & 29.4 & 28.0 & 33.2 & 26.0 & 19.2 \\
\ \ \ \ \ \ \ \  \ \ \ \  +GRPO & 32.9 & 26.9 & 29.2 & 26.5 & 17.0 \\
\bottomrule
\end{tabular}
}
\caption{Performance comparison between PPO and GRPO on ChatR1-3B across datasets.}
\label{tab:ppo_grpo_comparison}
\end{table}

\header{PPO vs GRPO.} While ChatR1 is trained with PPO, we provide in Figure~\ref{fig:train} training curves of \chatrone with Group Relative Policy Optimization (GRPO)~\cite{shao2024deepseekmath}, on both TopiOCQA and QReCC. Overall we observe that PPO is more stable than GRPO across datasets. ChatR1 trained with GRPO usually collapses after 100 or 200 steps, while PPO can be trained for longer. This aligns with previous work~\cite{jin2025empirical} on the stability comparison of GRPO with PPO. Table~\ref{tab:ppo_grpo_comparison} further reports the performance comparison of models trained with both optimizations. Despite the lower stability of GRPO, we observe that on two datasets, GRPO leads to better final performance.

\begin{table}[t]
\centering
\small
\adjustbox{max width=0.49\textwidth}{
\begin{tabular}{lcc|cc}
\toprule
 & \multicolumn{2}{c}{{\textbf{Training Signal}}} & \multicolumn{2}{c}{{\textbf{Response Generation}}} \\
\cmidrule(r){2-3} \cmidrule(l){4-5}
{Model} & {\textbf{Intent/Rewrite}} & {\textbf{N@3}} & {\textbf{F1}} & {\textbf{BERTScore}} \\
\midrule
ChatR1 & T5        & 22.2 & 24.7 & 71.7 \\
ChatR1 & Mistral   & 26.0 & 27.7 & 77.6 \\
ChatR1 & GPT-4.1   & 30.6 & 29.4 & 80.9 \\
\bottomrule
\end{tabular}
}
\caption{ChatR1 (3B) trained from silver quality intent reward, and impact on the downstream generation performance on TopiOCQA.}
\label{tab:intent_quality_vs_performance}
\end{table}

\header{Training intent quality.}
ChatR1 leverages human query rewriting as a user intent reward, to explicitly express the user's information need. While several datasets contain human rewrite, some such as TopiOCQA do not. Table~\ref{tab:intent_quality_vs_performance} compares using pseudo intent/rewrite labels from T5, Mistral, and GPT-4.1 when training ChatR1.
Overall, the performance follows the same pattern: better rewrites lead to higher F1 and BERTScore, while weaker signals hurt performance.

\begin{table}[t]
\centering
\small
\begin{tabular}{lcc}
\toprule
{Retrieval Method} & {\textbf{TopiOCQA}} & {\textbf{QReCC}} \\
\midrule
ChatR1-3B +Dense            & 29.4 & 28.0 \\
ChatR1-3B +BM25             & 17.6 & 26.6 \\
ChatR1-3B +Dense-Rerank   &  31.8 & 29.9 \\
\midrule
ChatR1-7B +Dense            & 30.6 & 31.0 \\
ChatR1-7B +BM25             & 22.8 & 29.4 \\
ChatR1-7B +Dense-Rerank   & 32.1 & 33.1 \\
\bottomrule
\end{tabular}
\caption{Generalization of ChatR1 to different retrieval systems during inference (F1-Score).}
\label{tab:retrieval_ablation}
\end{table}

\header{Retrieval tool.}
Table~\ref{tab:retrieval_ablation} compares ChatR1-3B and 7B under different retrieval setups at inference time. Models are trained with a dense retriever and evaluated with BM25, dense retrieval, and dense retrieval with reranking.
We observe that BM25 leads to a clear drop in performance, especially for the 3B model. In contrast, adding reranking on top of dense retrieval consistently improves results across both datasets. Interestingly, the 7B model is also more robust with smaller degradation when switching to BM25 (-11.8 F1 for 3B while -7.8 F1 for 7B). We also see that models trained on QReCC have much lower drop from using BM25 on QReCC compared to TopiOCQA, which can be explained as the retrieval task on QReCC is easier than on TopiOCQA.

\section{Conclusion}
\label{sec:conclusion}

We presented \chatrone, an RL-based reasoning framework for CQA, where user intent evolves across turns and must be inferred from context. Unlike static SFT pipelines that separate rewriting, retrieval, and generation, \chatrone jointly optimizes these steps end-to-end.
To address the sparsity of the outcome reward, we introduced an intent-aware reward component that aligns retrieval and reasoning with evolving user intents. Our experiments show that this design significantly improves both in-domain and out-of-domain performance over SFT baselines, while our analysis reveals adaptive reasoning behaviors such as context-aware query reformulation and variable reasoning depth.
Future work will explore dialogue-level optimization with simulated users and preference-based feedback, advancing RL-based reasoning for interactive information-seeking tasks.

\section*{Acknowledgements}

This research was partly supported by the Swiss National Science Foundation (SNSF), grant number 215742, on Personality and Conversational Information Access (PACINO).

\section{Limitations}

While \chatrone demonstrates that RL can improve reasoning in CQA, several aspects remain open for future work. Our framework relies on PPO/GRPO; exploring alternative optimization strategies, such as off-policy methods, or curriculum-based training could improve stability and sample efficiency. Current experiments focus on dialogues of moderate length (10–20 turns), whereas real-world interactions can be longer and require stronger memory and context modeling. Although smaller models already learn useful behaviors under RL, scaling to larger backbones may reveal stronger emergent reasoning and generalization. We also did not consider personalization or user-specific adaptation, which would be essential for more proactive and mixed-initiative dialogue. Finally, RL training introduces additional computational cost, both at training and inference; developing more efficient optimization schedules remains an important direction.




\bibliography{main}

\appendix

\newpage

\section{Prompts}
\label{sec:appendix:prompt}

We provide the prompts we used in this paper below. The main prompt used during the training of \chatrone is provided in the main paper, in Table~\ref{tab:prompts1}. We provide here the LLM-as-judge prompt in Table~\ref{tab:llmjudge_prompt}, we follow existing evaluation literature~\cite{song2025r1,rau-etal-2024-bergen}. We also provide the prompt we used to generate query rewrite both with GPT-4.1 and Qwen in Table~\ref{tab:teacher_prompt}.

\begin{table}[t]
  \caption{Prompt used for LLM-as-judge evaluation.}
  \label{tab:llmjudge_prompt}
    \begin{lstlisting}
Given a Question and its Golden Answer, verify whether the Predicted Answer is correct. The prediction is correct if it fully aligns with the meaning and key information of the Golden Answer. Respond with True if the prediction is correct and False otherwise.

Question: {question}
Golden Answer: {golden_answer}
Predicted Answer: {predicted_answer}
    \end{lstlisting}
\end{table}
\begin{table}[t]
  \caption{Rewrite prompt.}
  \label{tab:teacher_prompt}
    \begin{lstlisting}
    # Instruction:
    You are given a multi-turn conversation between a user and a system.
    Rewrite the last user question into a fully self-contained, context-independent query.
    # Guidelines:
    - Include all necessary information from previous turns.
    - Resolve pronouns and vague references into explicit entities.
    - Do NOT add information not mentioned in the conversation.
    - Be concise and natural.
    - Return only the rewritten query.
    
    # Conversation:{ctx}
    # Last user question:{user_utterance}
    # Rewritten query:
    \end{lstlisting}
\end{table}


\section{Datasets Statistics}
\label{sec:appendix:stat}
We provide in Table~\ref{tab:dataset_stats} the size of the five datasets we used, and their training/test splits. All datasets are composed of conversations with several turns. Datasets like TopiOCQA have long user-system interactions, with an average of more than 10 turns, while FaithDial usually has shorter interactions of around 5 turns. A turn here is defined as a pair (q,y) of both the user question and system response, making the history grow significantly.

\begin{table}[t!]
    \centering
    \small
    \adjustbox{max width=\columnwidth}{
    \begin{tabular}{lccccc}
    \toprule
    \multirow{2}{*}{Dataset} & \multirow{2}{*}{\textbf{\# Conv.}} & \multirow{2}{*}{\textbf{\# Turns}} & \textbf{Answer} & \textbf{Corpus} \\
    & & & \textbf{length} & \textbf{size} \\
    \midrule
    \multirow{2}{*}{TopiOCQA}  
        & \phantom{0}3,509 & 45,450 & 10.8 & \multirow{2}{*}{\phantom{0}25M} \\
        & \phantom{00,}205 & \phantom{0}2,514 & 11.2 & \\
    \midrule
    \multirow{2}{*}{QReCC}  
        & 10,823 & 63,501 & 16.8 & \multirow{2}{*}{\phantom{0}54M} \\
        & \phantom{0}2,775 & 16,451 & 17.6 & \\
    \midrule
    \multirow{2}{*}{INSCIT}  
        & \phantom{00,}249 & \phantom{0}1,443 & 30.9 & \multirow{2}{*}{\phantom{0}49M} \\
        & \phantom{00,}468 & \phantom{0}2,767 & 40.3 & \\
    \midrule
    \multirow{2}{*}{MultiDoc2Dial} 
        & \phantom{0}3,469 & 18,318 & 20.9 & \multirow{2}{*}{\phantom{0}10K} \\
        & \phantom{00,}660 & \phantom{0}3,266 & 20.0 & \\
    \midrule
    \multirow{2}{*}{FaithDial} 
        & \phantom{0}4,094 & 18,357 & 17.0 & \multirow{2}{*}{\phantom{0}21M} \\
        & \phantom{00,}791 & \phantom{0}3,539 & 17.6 & \\
    \bottomrule
    \end{tabular}
    }
    \caption{Statistics of the datasets used in our experiments. First and second rows correspond to train and test splits.}
    \label{tab:dataset_stats}
\end{table}

Since INSCIT contains both clarification turns and direct answer turns, we filtered out clarification turns. This reduces the test set from 2767 to 2197 turns. This dataset also contains turns with several gold answers for which we computed max F1 for a given turn on the two references. We filtered the training set in the same way, keeping only direct answers, and considered the multiple gold answer references as independent training samples, resulting in 1,844 samples.

We also include the corpus size. In particular, QReCC has a very large corpus, seven times larger than a dataset like MS MARCO commonly used in information retrieval. In contrast, MultiDoc2Dial is domain-specific, with a very dense passage collection on a sub-topic.

\section{Additional Analysis}

\begin{figure}[t!]
    \centering
    \includegraphics[width=\linewidth]{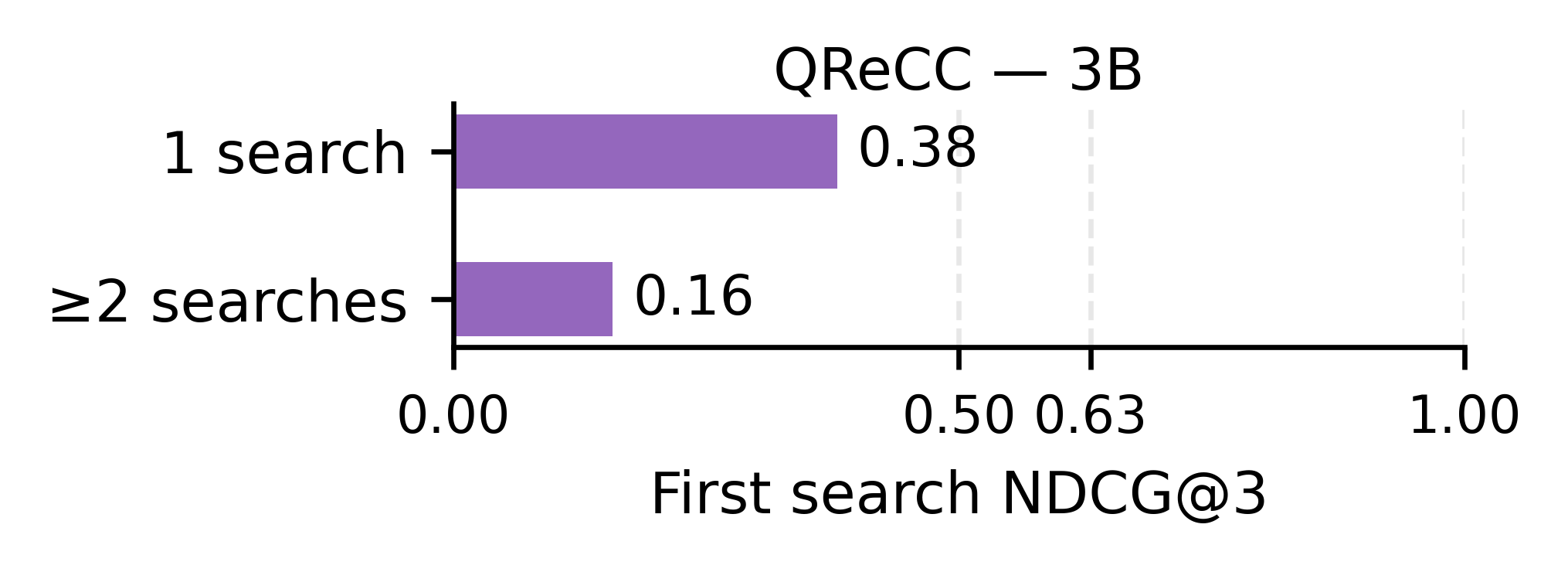}
    \caption{
    Performance of \chatrone first search, and decision. (1 search): \chatrone is satisfied with the retrieved passages and generates an answer. ($\geq$ 2 search): \chatrone was not satisfied with the first search, so it decided to perform another search.}
    \label{fig:firstsecond}
\end{figure}

Figure~\ref{fig:firstsecond} provides insight into when \chatrone decides to do another round of search. In particular, we observe that \chatrone decides to perform another search when the first search was not satisfactory. Internally, \chatrone thus has a threshold or its own judgment of whether the results are enough or if another search is needed.

\section{Baselines, and Experimental Details}
\label{sec:appendix:baselines}
The code to reproduce our experiments is available on the ChatR1 GitHub repository~\footnote{\url{https://github.com/SimonLupart/ChatR1}}, built upon the original Search-R1 repository~\cite{jin2025search} and veRL\footnote{\url{https://github.com/volcengine/verl}}. We also opensource datasets and model checkpoints on HuggingFace~\footnote{\url{https://huggingface.co/collections/slupart/chatr1}}.

We trained several \chatrone models, each with grounding from a different collection. When doing transfer and out-of-domain evaluation, we changed the base retrieval index to the new collection, both for baselines and \chatrone, making sure that relevant information can be found in the collection.

For the SFT baseline, we used a batch size of 512, a micro batch size of 16, a max input length of 1024, and a maximum generation length of 128 tokens. Learning rate is set to 1e-4, trained for 500 steps, with early stopping based on validation loss. This follows the setup of \chatrone.

We rely on \texttt{Qwen2.5-3B-Instruct} and its 7B variant as the base LLMs, fine-tuned on each five datasets. Retrieval model is \texttt{intfloat/e5-base-v2} used in zero-shot. \chatrone and other baselines use the top-3 retrieved passages; we also limit the number of search calls to two, following previous works~\cite{jin2025search,jin2025empirical}. Query rewrites are human annotations; when not available we used GPT-4.1 to generate them. The batch size of the policy model is 512, with a PPO micro batch size of 64, a max prompt length of 3500 tokens, and a learning rate of 1e-6 for the actor model. We rely on the GAE algorithm for the critic optimization. Both policy and critic models are initialized with the same LLM, with weights fine-tuned independently. We train for 500 steps, with a saved checkpoint every 50 training steps, taking the last checkpoint by default except if training rewards collapse. SFT models are also trained in the same fashion. All experiments are conducted on 4 H100 GPUs.

\section{Proximal Policy Optimization (PPO)}
\label{sec:appendix:ppo}
We train \chatrone with PPO, using the actor as policy model and critic as surrogate model. Previous works demonstrated that PPO was more stable than GRPO, hence we kept this formulation~\cite{jin2025search}. We also use a discount factor $\gamma$ and GAE parameter $\lambda$, both equal to $1$, following a previous study. In this setting, since the reward is only provided at the end of the trajectory, the advantage has a simplified formulation.
\begin{equation}
\hat{A}_i = R(\tau) - V_\psi(\tau_i)
\end{equation}
where $V_\psi$ is the learned value function (critic) and $R(\tau)$ the reward containing both intent and answer components. With this formulation, PPO is closer to the REINFORCE algorithm used in RL.
For PPO, we also use a $\epsilon=0.2$ in the clip loss objective.

We further provide the loss of the critic, trained with GAE in our case scenario.
\begin{equation}
\mathcal{L}_V(\psi)
=
\frac{1}{2}\,
\mathbb{E}_{(q,\mathcal{H};\mathcal{R};i)\sim\mu}
\Big[
\big(
V_\psi(\tau_i)
-
R(\tau)
\big)^2
\Big]
\end{equation}

\section{Qualitative Examples}
\label{sec:appendix:example}

We provide below several qualitative examples from \chatrone.

\begin{table*}[t]
\centering
\begin{tabular}{@{}p{0.95\textwidth}@{}}
\toprule

\textbf{Conversation History (\tctx):} \\[4pt]
\textbf{User:} The most relaxing color to me is green, what about you? \\
\textbf{Assistant:} I don't really find any relaxing but emeralds are green due to their chromium content, do you find that interesting? \\[4pt]
\textbf{User:} That is pretty interesting, I have noticed a lot of animals are green too, why is that? \\
\textbf{Assistant:} Many other creatures would beg to differ, a lot of them have adapted to their green environments by taking on their own green color to their skin. \\[4pt]
\textbf{User:} Yeah that is true but I was thinking of birds and ocean creatures if any of them can see color. \\
\textbf{Assistant:} Well, I don't really know about those animals being able to see colors but green is found between yellow and blue on the visible spectrum. \\
\midrule
\textbf{User (\tlast):}Since green is so common they're from plants right? What makes the color green anyway? \\
\midrule

\textbf{Golden Answer:} \\[4pt]
Yes, chlorophyll is the largest contributor for green. Chlorophyll is a chemical plants use to turn sunlight into chemical energy. \\
\midrule

\textbf{Reasoning, Search, and Information:} \\[6pt]

{\tthink} The user is asking about the presence of green in nature and what causes the color green. {\tethink}\\[6pt]

{\tsearch} What makes the color green? {\tesearch}\\
{\tinfo} Green is the color between blue and yellow on the visible spectrum. [...]{\teinfo}\\[6pt]

{\tsearch} Why is green common in nature? {\tesearch}\\
{\tinfo} Green is common in nature, especially in plants. Many plants are green mainly because of a complex chemical known as chlorophyll which is involved in photosynthesis. [...]{\teinfo}\\[6pt]

{\tthink} From the retrieved information, green’s prevalence in nature is mainly due to chlorophyll’s role in photosynthesis. {\tethink}\\
\midrule

\textbf{Predicted Answer:} \\[4pt]
{\tanswer} I found that the color green is common in nature primarily due to the presence of chlorophyll in plants, which is involved in photosynthesis. {\teanswer}\\
\midrule

\textbf{F1 Score:} 0.300 \\
\bottomrule
\end{tabular}
\caption{FaithDial (\chatrone 7B) -- Full reasoning with \tthink{}, showing how the model refines its second search query to then generate a correct answer.}
\end{table*}

\begin{table*}[t]
\centering
\begin{tabular}{@{}p{0.95\textwidth}@{}}
\toprule

\textbf{Conversation History (\tctx):} \\[4pt]
\textbf{User:} Why would I need to provide required verification for my FAFSA application? \\
\textbf{Assistant:} If you have received a note on your SAR saying you have been selected for verification or your school contacted you then you just have to confirm that the data reported on the FAFSA form is accurate. Have you been selected for verification? \\[4pt]
\textbf{User:} I have not been notified that I need verification. \\
\textbf{Assistant:} Unfortunately, no relevant information is found. \\[4pt]
\textbf{User:} What if I decide I want to get a loan to pay back? \\
\textbf{Assistant:} First, decide what aid you really need, and then respond to the school aid offer or choose to get a loan to pay back. \\
\midrule

\textbf{User (\tlast):}
Okay. Let's talk about Federal Student Aid renewal. In a presumed scenario, how will I handle the renewal? \\
\midrule

\textbf{Golden Answer:} \\[4pt]
No problem. Just Login into your account and select FAFSA renewal once there. It all will develop automatically. \\
\midrule

\textbf{Reasoning, Search, and Information:} \\[6pt]

{\tthink} The user is asking about the process of renewing Federal Student Aid (FAFSA). Typically, students need to reapply for financial aid each year, and there is a renewal process. {\tethink}\\[6pt]

{\tsearch} FAFSA renewal process {\tesearch}\\
{\tinfo} When you fill out the FAFSA form, you are applying for aid for a specific year. In order to receive aid the next year, you’ll need to submit that next year’s FAFSA form. [...]{\teinfo}\\[6pt]

{\tsearch} how will I handle the renewal of Federal Student Aid {\tesearch}\\
{\tinfo} If you filled out a FAFSA form last year and want to renew it, select LOG IN on the FAFSA home page, select I am the student, enter your FSA ID, and be sure to select FAFSA RENEWAL once given the option. [...]{\teinfo}\\[6pt]

{\tthink} From the information, it is clear that to renew the FAFSA, you need to log in and select FAFSA RENEWAL. {\tethink}\\
\midrule

\textbf{Predicted Answer:} \\[4pt]
{\tanswer} To renew your FAFSA, you just need to log in and select FAFSA RENEWAL. This will prefill many of the nonfinancial questions for you. {\teanswer}\\
\midrule

\textbf{F1 Score:} 0.342 \\
\bottomrule
\end{tabular}
\caption{MD2D (\chatrone 7B) -- Reasoning on domain-specific user question, requiring clear grounding from the corpus collection.}
\end{table*}

\begin{table*}[t]
\centering
\begin{tabular}{@{}p{0.95\textwidth}@{}}
\toprule

\textbf{Conversation History (\tctx):} \\[4pt]
\textbf{User:} What was the song \emph{The Night Chicago Died} about? \\
\textbf{Assistant:} In the song \emph{The Night Chicago Died}, the narrator retells his mother's anguish while awaiting news of the fate of her husband, a Chicago policeman. \\[4pt]
\textbf{User:} Who is the song \emph{The Night Chicago Died} by? \\
\textbf{Assistant:} \emph{The Night Chicago Died} is a song by the British group Paper Lace. \\
\midrule

\textbf{User (\tlast):} Does the group have other hits? \\
\midrule

\textbf{Golden Answer:} \\[4pt]
\emph{Billy Don't Be a Hero} is a 1974 pop song that was first a UK hit for Paper Lace. \\
\midrule

\textbf{Reasoning, Search, and Information:} \\[6pt]

{\tthink} Let's find out if the group Paper Lace has any other hits. 
=search> does the group Paper Lace have other hits? </search>  
My previous action is invalid. If I want to search, I should put the query between <search> and </search>.  
If I want to give the final answer, I should put the answer between <answer> and </answer>.  
Let me try again.
{\tethink}\\[6pt]

{\tsearch} does the group Paper Lace have other hits? {\tesearch}\\
{\tinfo} Their first two singles, released in 1974, were both written by Murray and Callander: “Billy Don’t Be a Hero” (No. 1 UK) and “The Night Chicago Died” (No. 1 US). [...] {\teinfo}\\[6pt]

{\tthink} Paper Lace had a hit with "Billy Don't Be a Hero". {\tethink}\\
\midrule

\textbf{Predicted Answer:} \\[4pt]
{\tanswer} Paper Lace had a hit with "Billy Don't Be a Hero". {\teanswer}\\
\midrule

\textbf{F1 Score:} 0.56 \\
\bottomrule
\end{tabular}
\caption{QReCC (\chatrone 7B) -- Example showing iterative self-correction during inference. While the system had to correct itself during the reasoning path, it still produced a good answer to the user.}
\end{table*}

\begin{table*}[t]
\centering
\begin{tabular}{@{}p{0.95\textwidth}@{}}
\toprule

\textbf{Conversation History (\tctx):} \\[4pt]
\textbf{\textbf{User:}} What type of meat is consumed at Christmas dinner in British Colonies? \\
\textbf{\textbf{Assistant:}} In many former British colonies the meats served are similar to those used in English Christmas dinners, and often involve roast meats. The primary meat served at Christmas in the U.K. is turkey, though historically boar, peacock, swan, and goose have also been served. \\[4pt]
\textbf{\textbf{User:}} What are the food served at Christmas dinner in Japan? \\
\textbf{\textbf{Assistant:}} A successful advertising campaign in the 1970s made eating at KFC around Christmas a national custom. Its chicken meals are so popular during the season that stores take reservations months in advance. \\[4pt]
\textbf{\textbf{User:}} How about Christmas dinner in Lebanon? What kind of food do they eat? \\
\textbf{\textbf{Assistant:}} In Lebanon, roast turkey is the most common choice of meal. Roasted duck, Lebanese salad (Tabbouleh), and pastries, such as honey cake, are common. \\
\midrule

\textbf{Query (\tlast):} 
Now I want to know the Christmas dinner in Honduras! \\
\midrule

\textbf{Golden Answer:} \\[4pt]
Christmas dinner is served around midnight on the 24th of December, consisting of tamales, roast pork leg, accompanied by ``torrejas,'' for dessert, and eggnog. \\
\midrule

\textbf{Search and Information:} \\[6pt]

{\tsearch} Now I want to know the Christmas dinner tradition in Honduras? {\tesearch}\\
{\tinfo} The traditional Honduran Christmas dinner is served around midnight on the 24th of December. [...]{\teinfo}\\[6pt]

{\tsearch} What kind of food is typically eaten during Christmas dinner in Honduras? {\tesearch}\\
{\tinfo} In Honduras, tamales are traditionally eaten on Christmas Eve evening. [...]{\teinfo}\\
\midrule

\textbf{Predicted Answer:} \\[4pt]
{\tanswer} In Honduras, the traditional Christmas dinner is served around midnight on the 24th of December and typically consists of tamales, roast pork leg, accompanied by "torrejas," for dessert, and eggnog. {\teanswer}\\
\midrule

\textbf{F1 Score:} 0.8627 \\
\bottomrule
\end{tabular}
\caption{INSCIT (\chatrone 3B) -- Example where the model infers from the context that the user is interested in examples of food, it thus refines the second search to better adapt to the user intent. We also see reduced thinking process here, reasoning is implicit, and made through the search queries.}
\end{table*}

\end{document}